\documentclass[lettersize,journal]{IEEEtran}
\usepackage{amsthm} 
\usepackage{algorithmic}
\usepackage{algorithm}
\usepackage{array}
\usepackage{textcomp}
\usepackage{stfloats}
\usepackage{url}
\usepackage{verbatim}
\usepackage{graphicx}
\usepackage{cite}
\usepackage{arydshln}
\usepackage{mathtools}
\usepackage{hyperref}
\usepackage{esint}
\usepackage{comment}
\usepackage{threeparttable}
\usepackage{xcolor}
\usepackage{latexsym}
\usepackage{multirow,bigdelim}
\usepackage{amssymb}
\usepackage{float}
\usepackage{amsmath}
\usepackage{esint}
\usepackage{mathrsfs}
\usepackage{subfigure}

\begin{document}

\title{Privacy at a Price: Exploring its Dual Impact on AI Fairness}

\author{ Mengmeng Yang, Ming Ding, Youyang Qu, Wei Ni, David Smith, Thierry Rakotoarivelo
\thanks{.}

}


\maketitle

\begin{abstract}
The worldwide adoption of machine learning (ML) and deep learning models, 
particularly in critical sectors, 
such as healthcare and finance,
presents substantial challenges in maintaining individual privacy and fairness. 
These two elements are vital to a trustworthy environment for learning systems. 
While numerous studies have concentrated on protecting individual privacy through differential privacy (DP) mechanisms, 
emerging research indicates that differential privacy in machine learning models can unequally impact separate demographic subgroups regarding prediction accuracy. 
This leads to a fairness concern, 
and manifests as biased performance. 
Although the prevailing view is that enhancing privacy intensifies fairness disparities, 
a smaller, yet significant, subset of research suggests the opposite view. 
In this article, 
with extensive evaluation results, 
we demonstrate that the impact of differential privacy on fairness is \emph{not} monotonous. 
Instead, 
we observe that the accuracy disparity initially grows as more DP noise (enhanced privacy) is added to the ML process,
but subsequently diminishes at higher privacy levels with even more noise. 
Moreover, 
implementing gradient clipping in the differentially private stochastic gradient descent ML method can mitigate the negative impact of DP noise on fairness.
This mitigation is achieved by moderating the disparity growth through a lower clipping threshold.

\end{abstract}

\begin{IEEEkeywords}
Differential privacy, fairness, the interplay between privacy and fairness
\end{IEEEkeywords}

\section{Introduction} 
In the era of big data and rapid technological advancement, 
machine learning, 
as an important component of AI, 
has become integral to our daily lives. 
Their applications span finance, healthcare, education, and criminal justice.
While these technologies provide convenient, efficient, and innovative solutions, they heavily depend on extensive personal datasets. 
Thus, these data-driven advancements raise significant privacy concerns, 
as demonstrated by incidents of sensitive medical records being inadvertently exposed and sophisticated adversarial attacks compromising user privacy. 
Privacy regulations, 
including the EU AI Act~\cite{AIAct}, 
have been established to safeguard individual data privacy while extending their purview to ensure fairness in AI applications, 
ensuring that personal data is protected while promoting equitable and unbiased outcomes. 

Differential privacy, 
as a provable notion of privacy, 
has emerged as a leading solution for addressing potential privacy breaches and has found widespread adoption in machine learning. This protective paradigm secures individuals' sensitive data by introducing carefully calibrated noise into model training processes or outputs, thereby establishing a robust protective mechanism that safeguards individual data privacy and facilitates the extraction of valuable insights from extensive datasets. However, recent studies reveal that while ensuring privacy, it can impact the accuracy of model outputs differently across groups, raising fairness concerns. 

The disparity effect of differential privacy in machine learning has been well-documented in theoretical and empirical studies. 
However, the literature presents \emph{conflicting} observations regarding its actual impact. 
Specifically, 
recent studies~\cite{bagdasaryan2019differential,tran2021differentially0,zhu2022post} show that differential privacy leads to an increase in the accuracy gap between underrepresented and adequately-represented groups, 
and this gap grows wider as the privacy budget decreases. 
In contrast, 
according to~\cite{xu2019achieving}, 
enhancing the level of privacy results in reduced model discrimination. 
Despite variations in models and datasets used, 
These findings strongly motivate us to investigate the influence of differential privacy on group accuracy thoroughly. 

This article is the first to reveal that the impact of differential privacy on fairness in terms of accuracy disparity is \textbf{NOT} `monotonous.’ 
Our experiments with standard machine learning algorithms on various datasets demonstrate that the accuracy gap initially widens with noise levels (indicating stricter privacy requirements) before gradually diminishing. 

Intuitively, 
as noise levels increase during the learning process, 
training the underrepresented group, such as a minority class, 
with fewer samples becomes more challenging, 
resulting in a larger accuracy gap. 
Moreover, 
when noise is added to the model's output, 
it often misplaces samples across decision boundaries, 
especially in cases where the decision region is small, 
as often with minority or underrepresented groups. 
Consequently, 
the model's performance degrades more for these groups than well-represented groups, 
widening the accuracy disparity. 
Furthermore, 
as the noise level continues to increase and beyond a certain point, 
the noise becomes dominant, 
and excessive noise muddles the important signals the model needs to make informed predictions, 
leading to uniformly poor, 
albeit fair, 
performance across the groups.

We also observe that implementing clipping in the differentially private stochastic gradient descent (DP-SGD) ML method curbs the impact of differential privacy on the accuracy disparity. 
Notably, 
a smaller clipping bound requires a much stronger privacy level to widen the accuracy gap. 
The intuition behind this is that given the same level of privacy budget, 
the smaller clipping bound (sensitivity) reduces the noise introduced to the model. 
As a result, 
the accuracy gaps in scenarios with a smaller clipping bound increase slower than those with larger ones.

In conclusion, 
our research provides a novel perspective and insights into the impact of differential privacy on fairness. 
We aim to enhance the understanding of this issue and stimulate deeper consideration when conducting private data analysis and developing AI applications using differential privacy.

\section{Background}

In this section, we illustrate existing research on privacy and its associated fairness issues.

\subsection{Differential privacy}
Differential privacy was proposed by Dwork et al.~\cite{dwork2006differential} in 2006. 
As a provable privacy notation, it has arisen as a paradigm choice for preserving privacy in a broad spectrum of applications. 
It ensures that adding, removing, or altering an individual's record within a dataset does not substantially impact the overall statistical results. 
Differential privacy is achieved by introducing controlled randomness into data, making it exceedingly challenging to discern any specific individual's data while ensuring statistical accuracy. 

Numerous differential privacy mechanisms have been proposed. The typical ones include the Laplace mechanism, the Gaussian mechanism, and the Exponential mechanism. 
Specifically, the Laplace Mechanism adds Laplace-distributed noise to query results, while the Gaussian Mechanism uses Gaussian-distributed noise. Both mechanisms determine noise scales based on the query's sensitivity and a privacy parameter $\epsilon$, with smaller $\epsilon$ values providing stronger privacy but potentially noisier results. On the other hand, the Exponential Mechanism is used to choose specific items from a dataset while preserving differential privacy. The method calculates scores for possible options using sensitivity and $\epsilon$. It then selects items probabilistically and introduces randomness to ensure privacy is preserved.

This article explores two distinct perturbation approaches, as illustrated below. \\
\noindent \textbf{Output perturbation.} Output perturbation refers to the privacy-preserving method that protects the sensitive information in the output of computations or algorithms \cite{dwork2006calibrating}. It adds controlled random noise or perturbation to the final output, such as the model predictions, to obfuscate the exact values.  \\
\noindent \textbf{Gradient perturbation.} Gradient perturbation works by introducing small random noise or perturbations to the gradients of a model's parameters during training. The most popular and extensively used gradient perturbation framework is Differentially Private Stochastic Gradient Descent (DP-SGD)~\cite{abadi2016deep}, which works by injecting Gaussian noise to the clipped gradient to protect individual data while maintaining model utility. 

\subsection{Fairness}

Fairness in machine learning has been a key challenge because algorithms and models can be susceptible to bias and discrimination, particularly when trained on biased or incomplete data. This can lead to unequal treatment of different individuals or groups, perpetuate existing social disparities, and undermine trust in machine learning and AI systems. 

Fairness in machine learning can be approached from two main perspectives: \textit{individual fairness} and \textit{group fairness}. Individual fairness is the idea that similar individuals should be treated similarly. Group fairness, on the other hand, is the idea that different demographic groups should be treated fairly. For example, the model performance across different groups should be similar.  Both individual fairness and group fairness are important considerations in machine learning.  However, group fairness may be seen as a more holistic and comprehensive approach to ensure fairness and equity, considering the needs and experiences of multiple individuals with consistent characteristics or identities. In addition, group fairness is often easier to define and measure than individual fairness. The literature has tended to focus more on group fairness, while we also consider group fairness in this article. The following are commonly adopted group fairness notions:

\noindent\textbf{Demographic parity.} Demographic parity, in the context of fairness in machine learning, refers to a situation where for any group represented by $a\in A$ in a given dataset and for any possible prediction label $k\in Y$, the probability that the model predicts the label $k$ for a member of that group $a$ is equal to the overall probability that the model predicts the label $k$ for the entire dataset.

\noindent\textbf{Accuracy parity.} Accuracy parity is satisfied when the predictor's misclassification rate is conditionally independent of the sensitive group attribute. The misclassification rate within a specific group is identical to that across the entire dataset. 

\noindent\textbf{Equal odds.} The equalized odds criterion requires that the model's predicted outcomes be independent of the sensitive attribute, conditional on both true positive rates and false positive rates. 

\noindent\textbf{Equal opportunity.} Equal opportunity only requires that the true positive rates are equal across different subgroups defined by the sensitive attribute. 

\noindent\textbf{Excessive risk.} Excessive risk defines the differences between the private and non-private empirical risk function. 
Fairness is determined by the greatest discrepancy in excessive risk between a single group and the overall dataset. Fairness is achieved when there is no difference in excessive risk across all protected groups. 

Following \cite{bagdasaryan2019differential},  which was the first to identify the impact of differential privacy on disparity, we also employ the accuracy parity as our fairness metric.  We will explore other fairness metrics for our future work. 

\section{Interplay Between Privacy and Fairness}

Differential Privacy focuses on protecting the privacy of individuals by adding noise to the data or query responses so that it becomes impossible to distinguish the contribution of any particular individual to the output. In contrast, fairness aims to ensure that computation outputs are equitable and unbiased across different groups of individuals. 
differential privacy is a privacy protection technique, while fairness is concerned with mitigating discrimination or bias in data and algorithms. 
They are two distinct concepts. However, there is an important relationship between differential privacy and fairness. 

\subsection{Privacy degrades fairness}
Bagdasaryan et al.~\cite{bagdasaryan2019differential} were the first to show that differential privacy has a disparate impact on model accuracy. They observed that the reduction in accuracy caused by DP-SGD disproportionately affects under-represented groups, which in turn impacts the fairness of the model. They also show that the accuracy of private models tends to decrease more for classes that already have lower accuracy in the original, non-private model. 
In addition, they observed that noise addition is the main factor that impacts the accuracy of underrepresented subgroups. This is because these groups may have less data available for the model to learn from, and the added noise can make it more difficult for the model to learn from that limited data accurately. Xu et al.~\cite{xu2021removing} found that the gradient clipping for DP-SGD also causes bias. When different groups significantly differ in the magnitude of the gradients, and these values exceed the clipping bound, the gradient clipping technique leads to unequal information losses among these groups, eventually penalizing the groups with greater gradients. 

The disparity effect of differential privacy has been observed on another differentially private learning mechanism~\cite{tran2023fairness}, named Private aggregation of Teacher Ensembles (PATE)~\cite{papernot2016semi}, but it causes milder disparate impacts when compared with DP-SGD~\cite{uniyal2021dp}, attributed to the teacher-ensemble setting. 
In addition to the model characteristics, Tran et al.~\cite{tran2023fairness}  showed that the properties of the training data affect the disparate impacts as well. For example, the input norm has a strong correlation with the expected model sensitivity and, in turn, the disparate impacts of the private model.

Pujol et al.~\cite{pujol2020fair} observed such a disproportionate effect of differential privacy on decision-making. Specifically, they empirically proved that several decision problems, such as allocating educational funds with high societal impact, induce inherent biases when using differential private input. Tran et al.~\cite{tran2021decision} found that the groups with a small distance to the decision boundary are less robust to noise in the model. Therefore, adding noise causes more errors in model decisions on these groups. The dissimilar error introduced to different groups exacerbates the unfairness. Zhu et al.~\cite{zhu2022post} showed that the post-processing of differential privacy also causes fairness issue; i.e., it inherently introduces unfairness when post-processing the noisy outcome using the projection method, if the true value is not at the centroid of the feasible region.

\subsection{Fairness increase privacy risk}
Measuring how fairness affects privacy, especially using differential privacy, is tricky because we can't easily quantify the privacy level without introducing any randomization operation. To check if a fairness-aware model impacts privacy, Chang and Shokri~\cite{chang2021privacy} analyze the information leakage of the model through membership inference attacks that infer whether a data point is used for training a model. 
Their empirical findings reveal a significant and unequal impact of fairness-aware learning on the privacy risks of subgroups, with a notable increase in privacy risk for the unprivileged subgroup (eg. the dataset is complex or the size of it is small). The higher the fairness achieved by a model, the more it tends to heighten privacy risks for this disadvantaged subgroup. The work shows that the fair model tends to memorize data from the underrepresented subgroups due to over-fitting while striving to balance the model's error across different groups. This memorization leads to an increase in the model’s information leakage about unprivileged groups. 

Research exploring the impact of fair learning on privacy is still in its early stages. We'll look into this topic in the future as part of our ongoing research. 

The relationship between privacy and fairness is more intricate than we previously understood. In the next section, we focus on exploring the impact of differential privacy on fairness in machine learning. This exploration is motivated by conflicting observations: while some literature suggests that differential privacy negatively impacts fairness, others indicate a positive impact.

\section{Evaluation}

To investigate fairness under privacy guarantees, we conduct output and gradient perturbations on various machine learning models using different datasets. We show how the accuracy disparity varies across different groups as the privacy level increases.

\subsection{Output perturbation} 
\begin{figure}[ht]
     \centering
\includegraphics[scale=0.55]{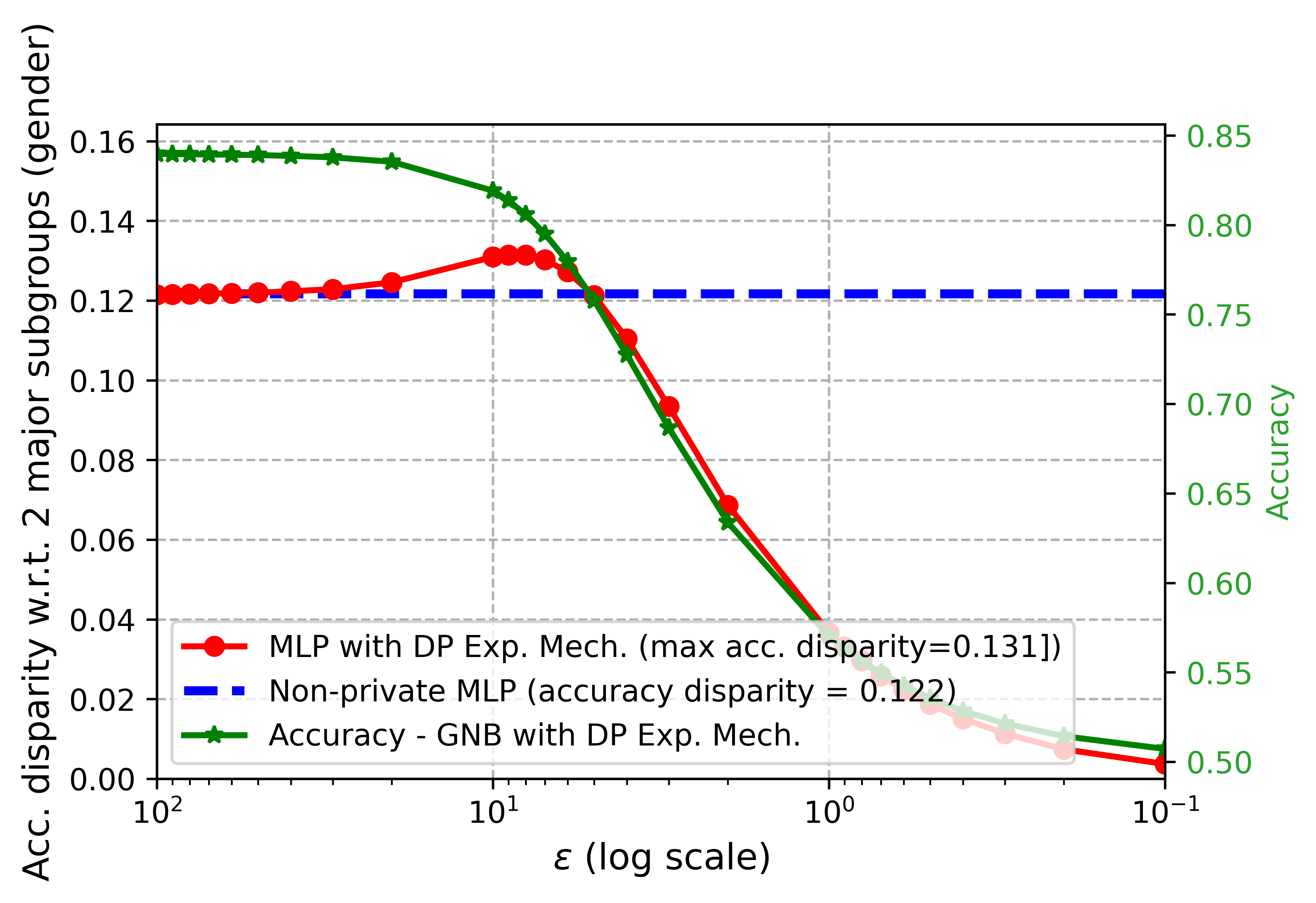}
\caption{Accuracy disparity on the Adult dataset (output perturbation)}
\label{fig_em_adult}
\end{figure}

\begin{figure}[ht]
     \centering
\includegraphics[scale=0.55]{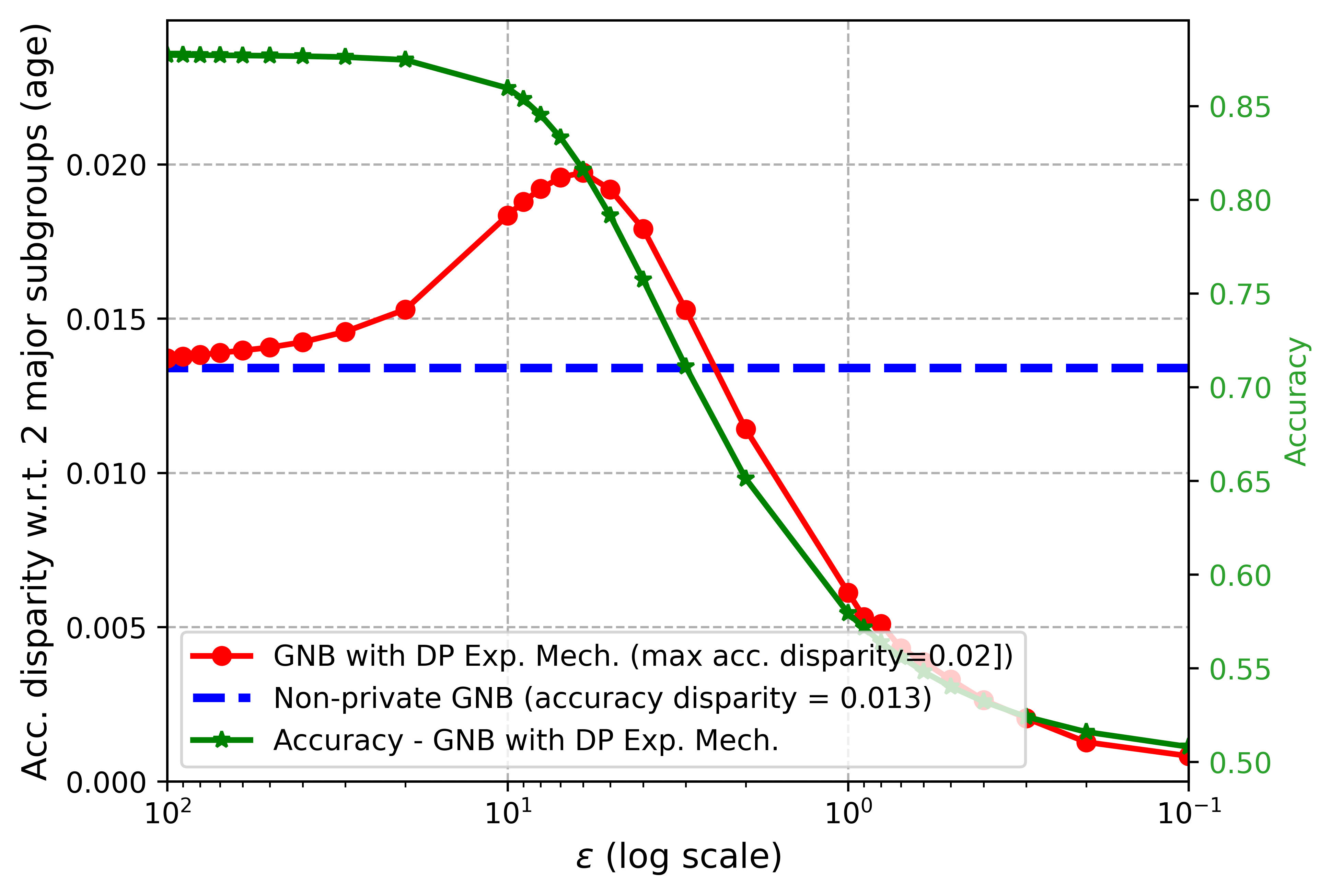}
\caption{Accuracy disparity on the Bank dataset (output perturbation)}
\label{fig_em_bank}
\end{figure}

In this Section, we achieve differential privacy by introducing randomization to the output of the machine learning. Specifically, we incorporate the Exponential Mechanism to obfuscate the predictions' output. \\
\noindent\textbf{Dataset.} To assess the impact of differential privacy with output perturbation, we conduct our analysis on two distinct datasets. The first dataset, known as the Adult dataset, comprises demographic details of individuals, with gender serving as the sensitive attribute for predicting income. The second dataset, referred to as the Bank dataset, provides information on marketing campaigns carried out by a Portuguese banking institution, where age is considered the sensitive attribute for predicting client subscription to a term deposit. \\
\noindent\textbf{Model.} We experimented using two distinct models. Specifically, we applied a Multi-Layer Perceptron (MLP) to the Adult dataset and employed Gaussian Naive Bayes (GNB) on the Bank dataset. For MLP, we utilize a neural network architecture consisting of two fully connected layers with $128$ units, a batch size of $512$, a ReLU activation function, and a cross-entropy loss function, and employ the Adam optimizer with a learning rate of $10^{-4}$. We used the GaussianNB classifier directly from the sci-kit-learn library for GNB. \\ 
\noindent\textbf{Setting.} We partition each dataset into a $70\%$ training set and a $30\%$ testing set for our experimentation. We vary the privacy budget $\epsilon$ in the range $[0.1, 100]$. We train all models over $30$ epochs, repeat it $10,000$, and evaluate the average. \\
\textbf{Results.} 
Figs.~\ref{fig_em_adult} and~\ref{fig_em_bank} depict the outcomes for the Adult and Bank datasets on MLP and GNB models, respectively. It's evident that when the privacy budget is sufficiently large, the accuracy disparity mirrors that of the model without differential privacy, which means a small amount of noise is not enough to make the disparity impact on the group accuracy. However, as the level of noise increases, the model's accuracy on the entire dataset decreases. The accuracy gap grows and it subsequently diminishes beyond a certain threshold, ultimately converging to $0$. This indicates that the accuracy for different groups becomes equal, alleviating fairness concerns. We can observe that maintaining an acceptable accuracy rate, such as $80\%$, a higher level of privacy may result in a fairer model, while still maintaining good accuracy. For instance, in the Bank dataset, at its peak, the model achieves an accuracy of around $82\%$. As $\epsilon$ becomes smaller, the accuracy does not decrease significantly, remaining above $80\%$, while the accuracy disparity is reduced, which result in a relatively more fair model.


\subsection{Gradient perturbation}
In this Section, we achieve differential privacy using DP-SGD. Rather than perturbing the model output, we introduce differential privacy noise to the gradients during the training process of the machine learning model. \\
\noindent\textbf{Dataset.} 
In addition to the Adult and Bank datasets, our experimentation extends to the MNIST dataset, a collection of grayscale images featuring handwritten digits from $0$ to $9$, with $60,000$ training samples and $10,000$ testing samples. In alignment with prior research such as~\cite{bagdasaryan2019differential}, we identify class $2$ as the well-represented class, characterized by the fewest false negatives, and class $8$ as the under-represented class with the highest false negatives. For class $8$, we limit the training samples to just $2,000$, reducing the total training samples to $56,149$. 

\noindent\textbf{Model.} 
We utilize both MLP and CNN models for gradient perturbation-based training.  For the MLP, we adopt the same settings as those used for output perturbation but integrate DP-SGD to guarantee privacy. We build the MLP on both the Adult and Bank datasets. As for the CNN model, which is used on the MNIST dataset, following the approach presented in~\cite{bagdasaryan2019differential}, we utilize two convolutional layers and two linear layers, collectively featuring $431,000$ parameters. 
Regarding hyperparameters, we adopt a batch size of $256$, a learning rate of $0.05$, and $60$ training epochs.

\begin{figure*}[ht]
     \centering
\subfigure{
\includegraphics[scale=0.35]{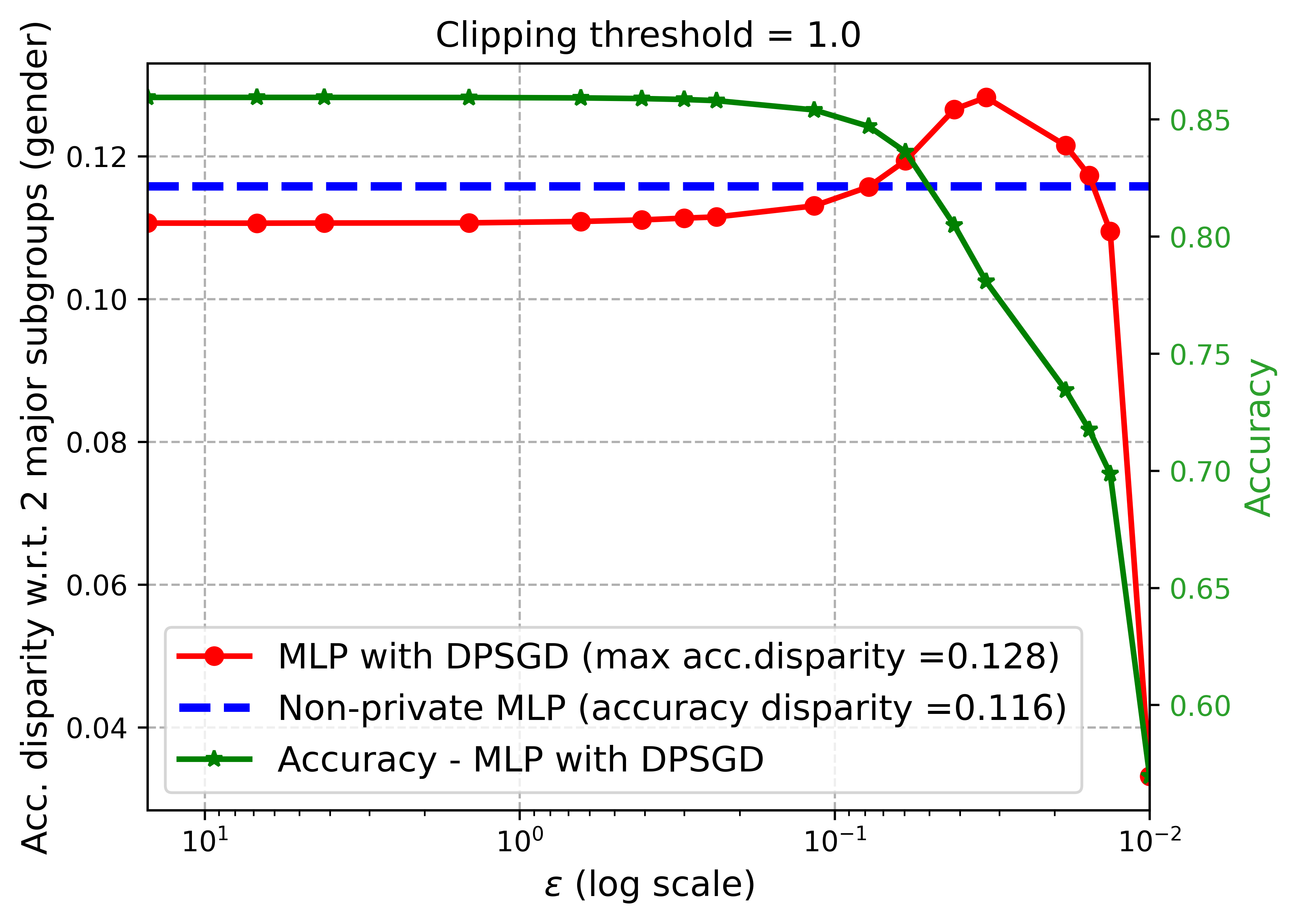}
}
\subfigure{
\includegraphics[scale=0.35]{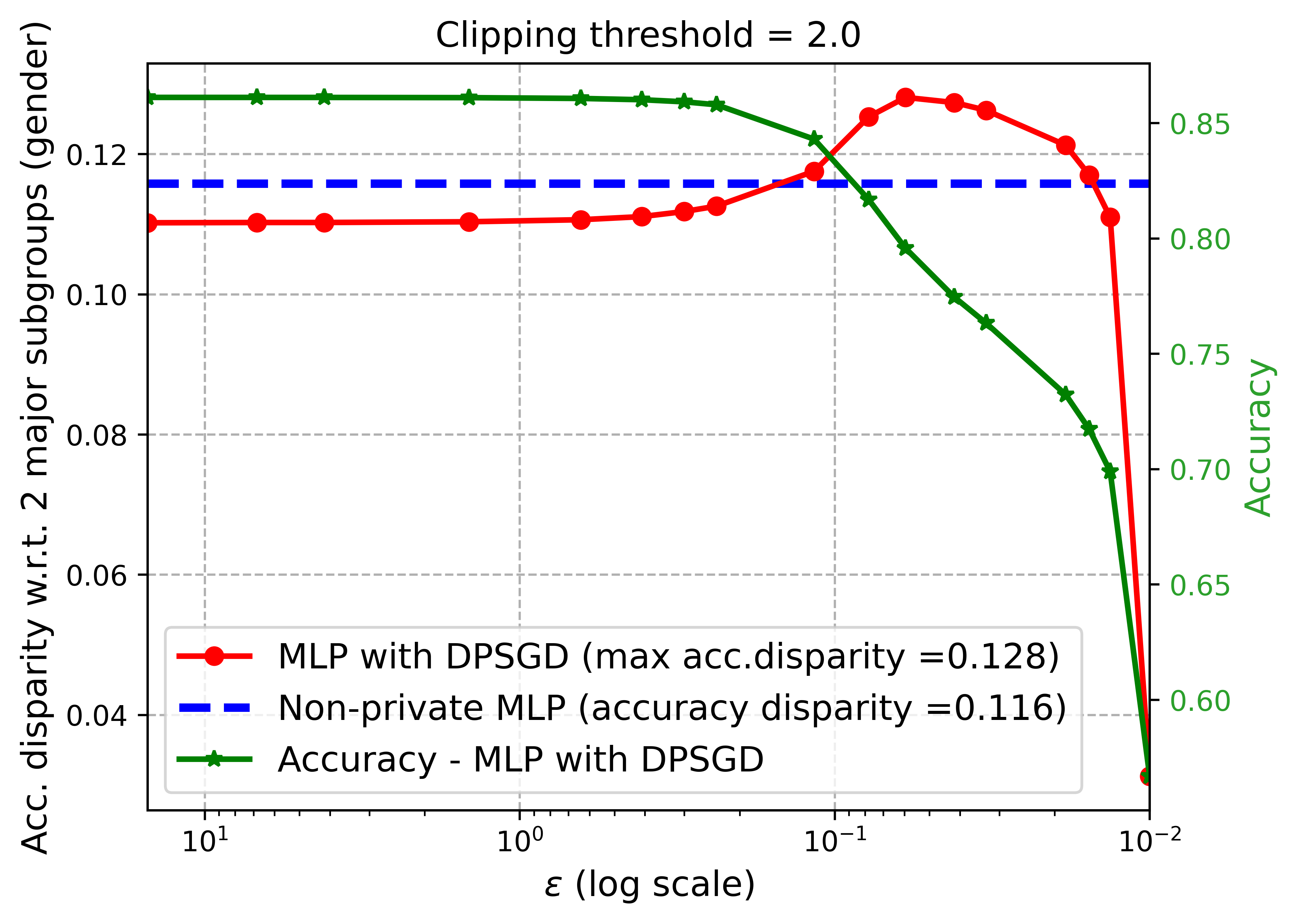}
}
\subfigure{
\includegraphics[scale=0.35]{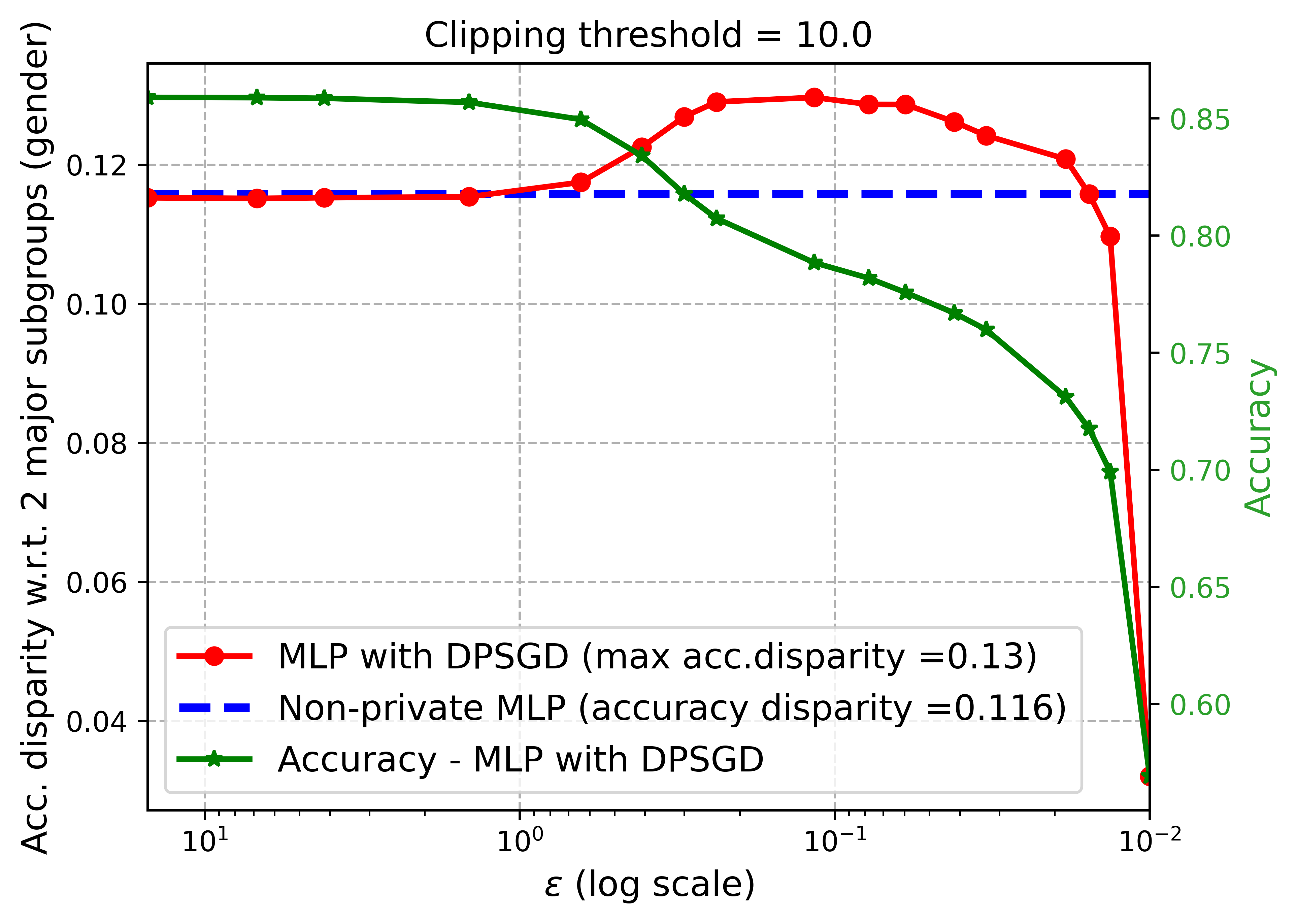}
}
\caption{Accuracy disparity on Adult dataset (gradient perturbation)}
\label{fig_dpsgd_adult}
\end{figure*}

\begin{figure*}[ht]
     \centering
\subfigure{
\includegraphics[scale=0.345]{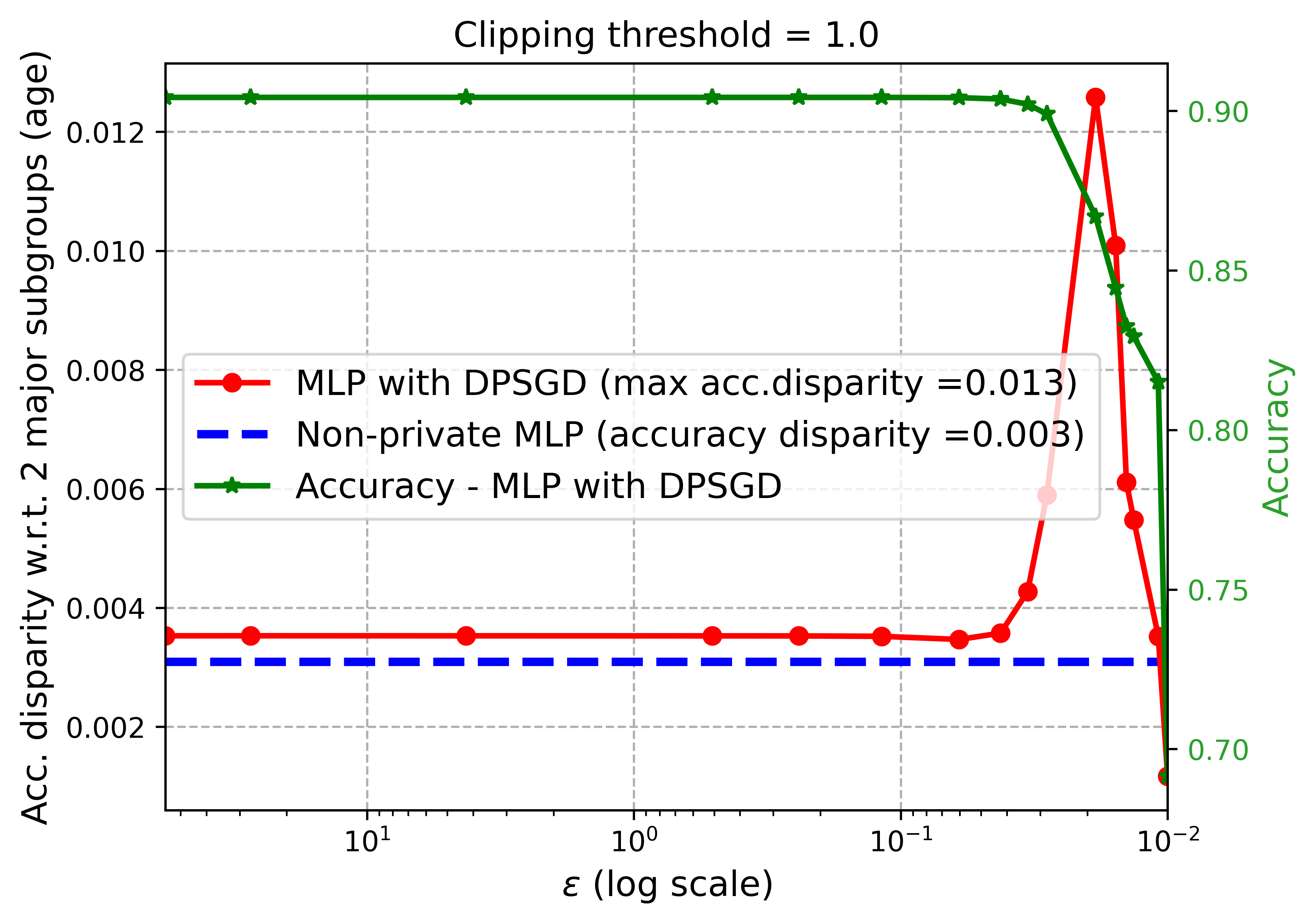}
}
\subfigure{
\includegraphics[scale=0.345]{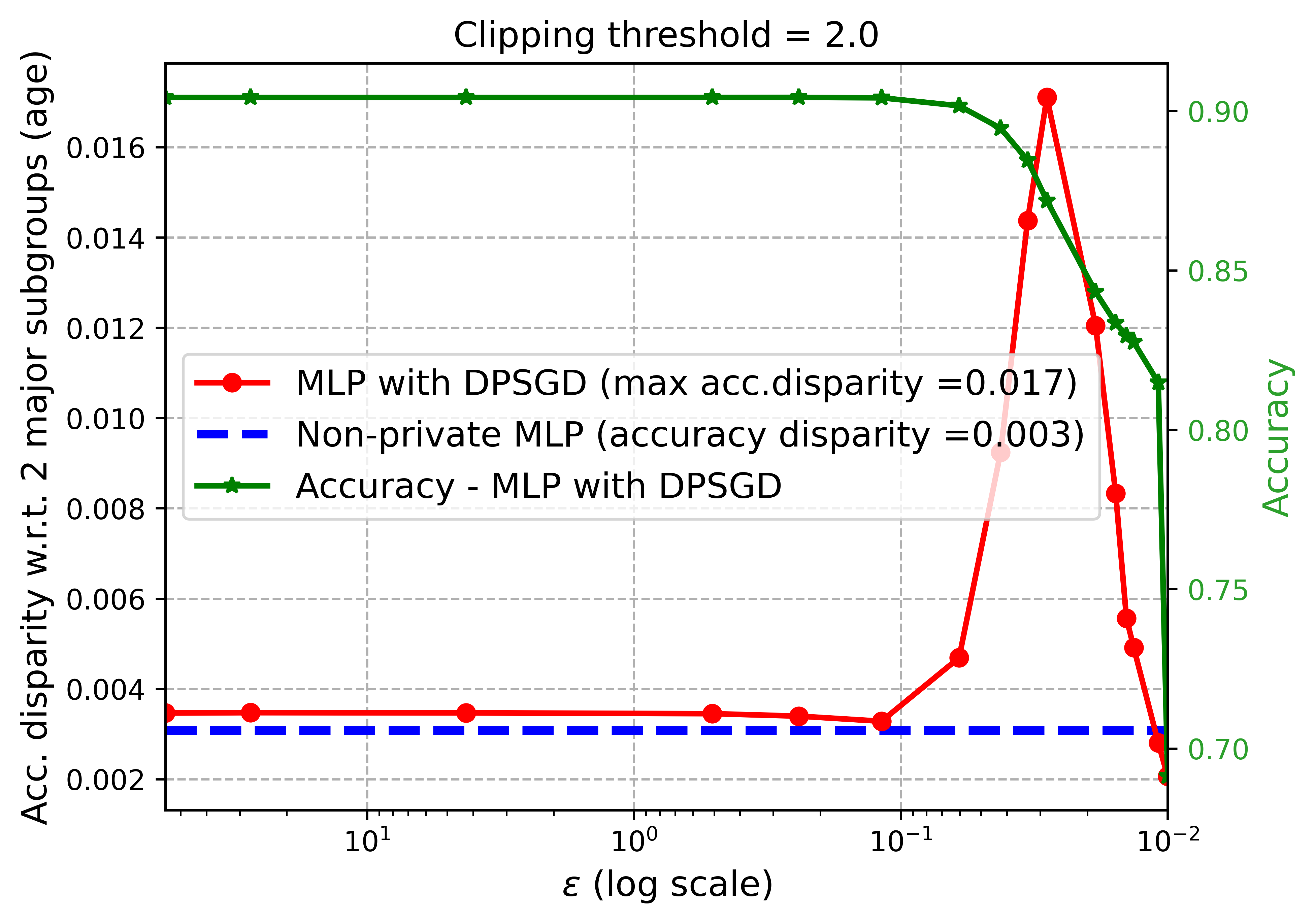}
}
\subfigure{
\includegraphics[scale=0.345]{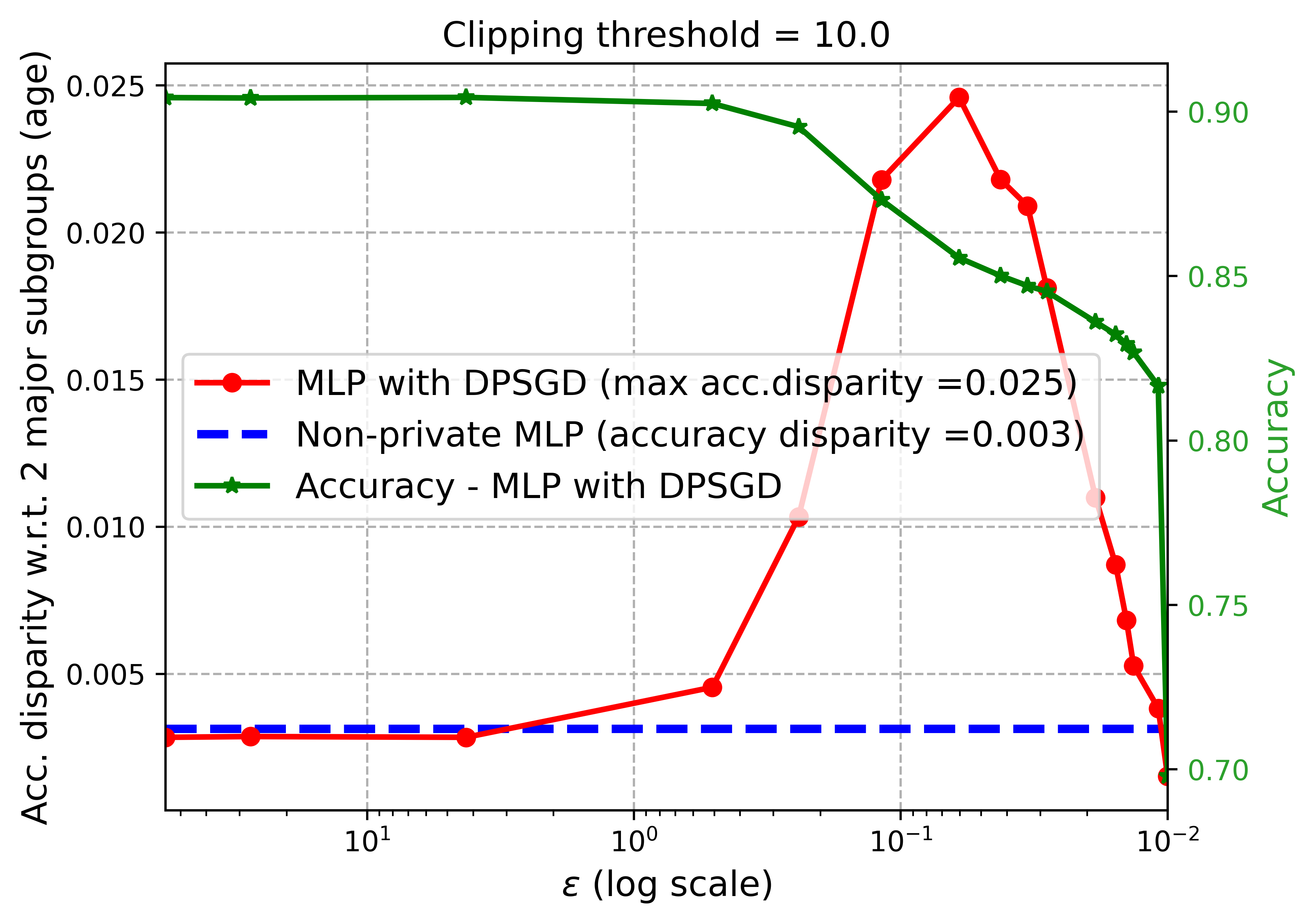}
}

\caption{Accuracy disparity on Bank dataset (gradient perturbation)}
\label{fig_dpsgd_bank}
\end{figure*}

\begin{figure*}[ht]
     \centering
\subfigure{
\includegraphics[scale=0.35]{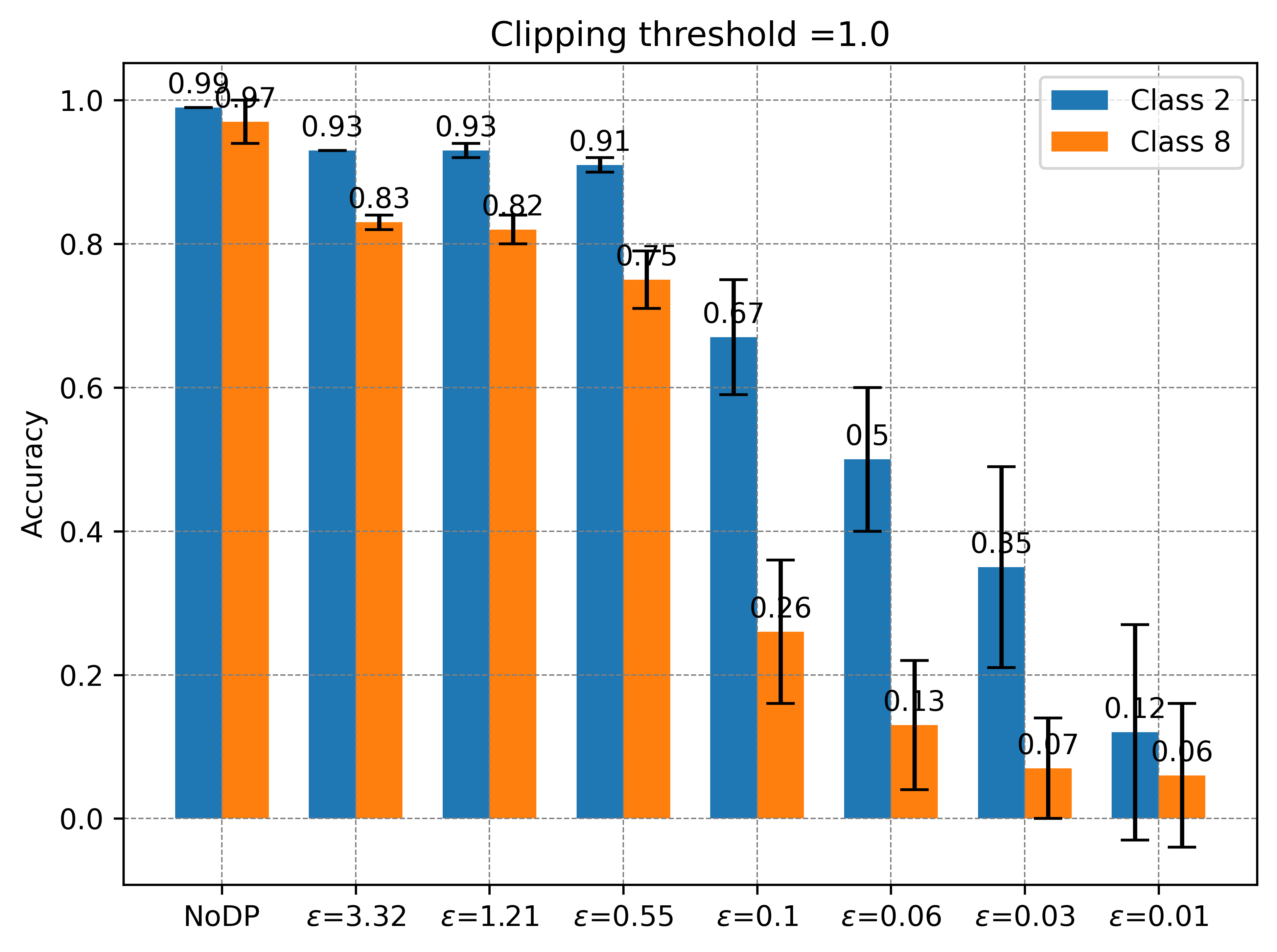}
}
\subfigure{
\includegraphics[scale=0.35]{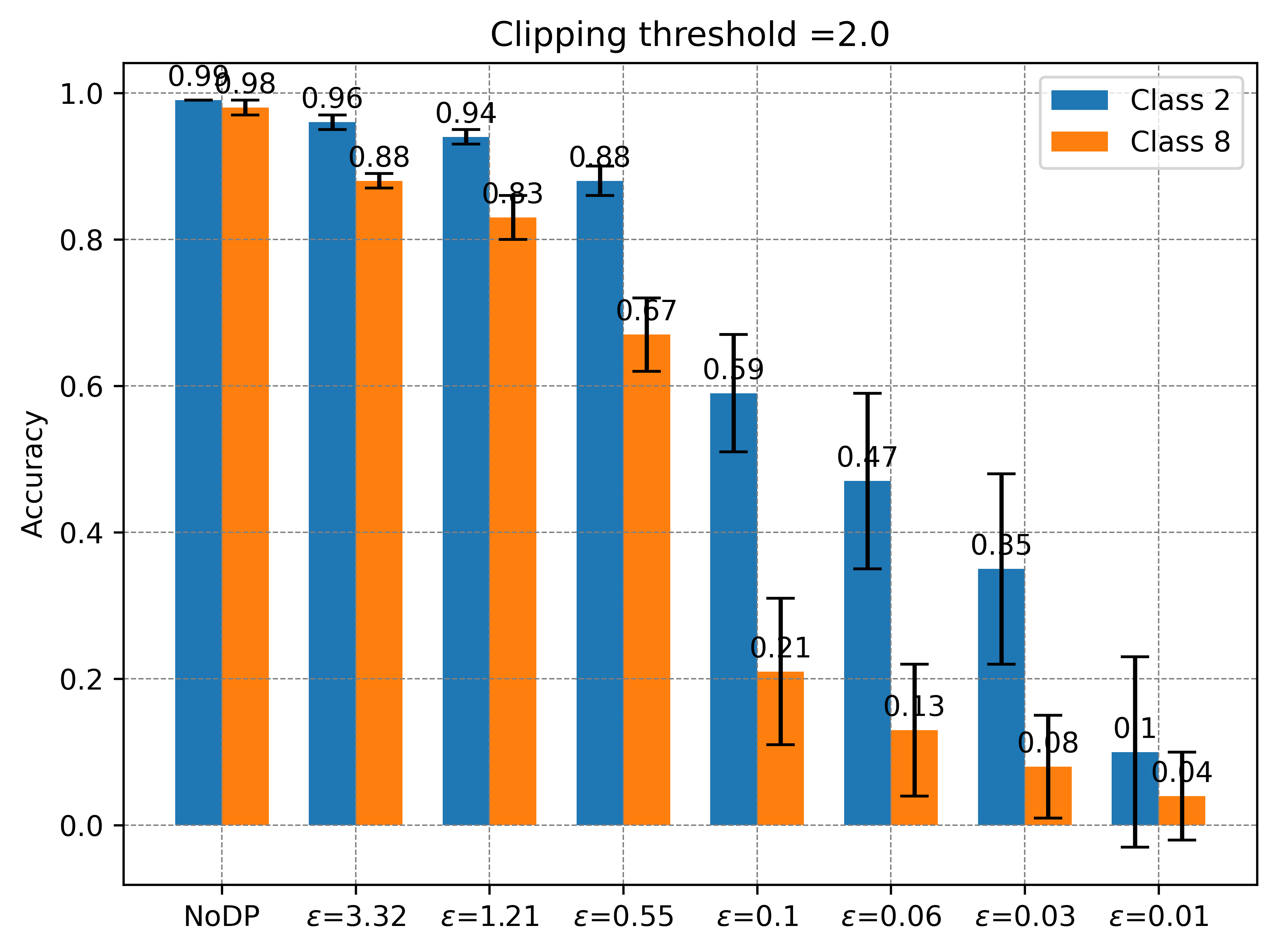}
}
\subfigure{
\includegraphics[scale=0.35]{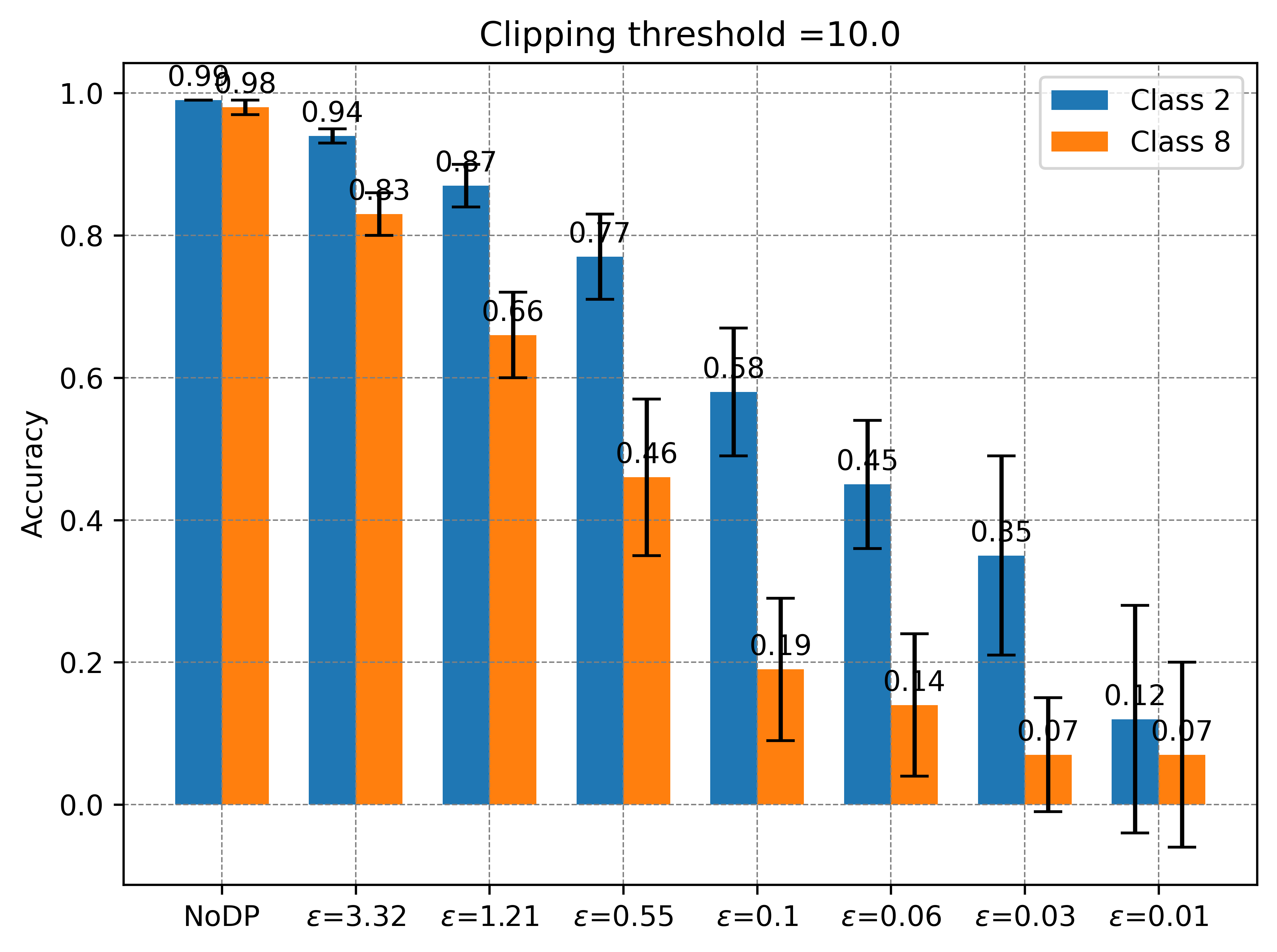}
}
\caption{Accuracy disparity on MNIST dataset (gradient perturbation)}
\label{fig_dpsgd_mnist}
\end{figure*}

\noindent\textbf{Setting.} We set the privacy parameter $\delta$ to a fixed value of $10^{-6}$, while varying $\sigma$ across the range from $0.5$ to $10,000$. We employ the Renyi Differential Privacy (RDP) accountant to monitor and track the privacy costs. 
To assess the impact of clipping bounds on accuracy disparity, we vary the clipping bounds, testing with values of $1$, $2$, and $10$. 
To ensure robust results, we repeat the training process for the MLP on both the Adult and Bank datasets at least $1,000$ times, averaging the outcomes. Similarly, we repeat the training for the CNN model $300$ times and take the average of the results. \\
\noindent\textbf{Result.} 
Fig.~\ref{fig_dpsgd_adult} presents the results for the Adult dataset. It is evident that the accuracy disparity does not simply increase or decrease with the privacy budget. It initially rises with the introduction of noise, then decreases after reaching a specific threshold, and becomes minimal when a substantial amount of noise is added. A significant amount of noise causes the model's predictions to resemble random guesses, leading to similar accuracy across different groups. 
Additionally, the impact of gradient clipping on accuracy disparity is noteworthy. With a sizable privacy budget and a clipping threshold set to $10$, the accuracy disparity under differential privacy closely resembles that of the model trained without differential privacy. However, when the clipping threshold is smaller, the accuracy disparity reduces.
Furthermore, we observe that the clipping threshold `postpones' the privacy level's impact on accuracy disparity. Specifically, when the clipping threshold is set to $10$, the accuracy disparity starts to increase at an epsilon of around $1$. As the clipping threshold increases to $2$, the point at which accuracy disparity begins to rise shifts to higher epsilon values, and with a clipping threshold of $1$, the accuracy disparity only starts to increase when $\epsilon$ is approximately $0.1$. Similar results can be observed on the Bank dataset shown in~Fig.~\ref{fig_dpsgd_bank} and the MNIST dataset shown in~Fig.~\ref{fig_dpsgd_mnist}. 
In addition, the green curve represents the accuracy of the model across the entire dataset. Analyzing these results provides us with valuable insights for parameter selection. As depicted in Fig. \ref{fig_dpsgd_bank}, when the clipping threshold is set to $10$, opting for $\epsilon=0.1$ provides a robust privacy guarantee. However, examining the accuracy disparity reveals a notable fairness issue at this threshold. Nonetheless, as the privacy level increases, the model maintains an accuracy of over $80\%$ while also demonstrating improved fairness across different groups. In such scenarios, opting for a higher privacy level may be preferable.


\section{Take-away}
Through the evaluation results obtained, we have two key findings as follows. 
\begin{itemize}
    \item \textbf{The impact of differential privacy on fairness is not ``monotonous''.} 
    Our empirical analysis reveals that the influence of differential privacy on fairness is not linear or monotonic as anticipated. 
    While the accuracy gap initially increases with rising privacy levels, 
    as most studies suggest, 
    it begins to diminish beyond a certain threshold of privacy level and eventually vanishes.
    Intuitively, 
    introducing noise into the learning process will impair the model's ability to extract accurate insights from underrepresented groups compared to well-represented ones, 
    leading to an AI accuracy gap for different groups. 
    When differential privacy is applied by adding noise to a machine learning model's output, 
    it disrupts the model's ability to grasp subtle data patterns, 
    notably those linked to smaller or underrepresented subgroups. 
    This noise tends to displace samples across decision boundaries, 
    notably harming the performance of minority groups compared to well-represented ones, 
    widening the accuracy disparity.
    On the other hand, 
    at an exceedingly high privacy level, 
    excessive noise can overwhelm the original data, 
    degenerating the model's predictions into random guesses.   
    In this extreme case, 
    prediction accuracy becomes uniform across all data records, 
    irrespective of group association, 
    eliminating any accuracy gap or fairness concern. 
    
    \item \textbf{The gradient clipping alleviates the disparity issue caused by the DP noise.} 
    Our empirical evidence has established that gradient clipping is pivotal in influencing accuracy disparity. 
    As corroborated by other research, 
    this effect can either positively mitigate or negatively exacerbate the disparity, 
    with data and model characteristics playing a determining role. 
    A deeper investigation is necessary to precisely delineate how gradient clipping affects this disparity.     
    Notably, 
    our research is the first to illuminate that a smaller clipping threshold can effectively `postpone' the impact of DP noise on accuracy disparity. 
    This is because, 
    under a constant privacy budget, 
    a smaller clipping threshold leads to reduced sensitivity, 
    thereby diminishing the noise introduced into the model. 
    As a result, 
    for a given privacy level, 
    a model with a smaller clipping threshold incurs less noise. 
    Consequently, 
    a more robust privacy level is required to replicate the same accuracy gap observed in models trained with a larger clipping threshold.

\end{itemize}

\section{Conclusion}

Privacy and fairness are integral pillars of trustworthy AI. 
It's crucial to ensure both aspects when applying machine learning or deep learning models to various domains, 
especially under the lens of recent AI regulations. 
In our study, 
we undertook empirical research to scrutinize the impact of differential privacy on accuracy disparity. 
Our results reveal a significant discovery: 
the accuracy gap widens as privacy levels rise, 
but intriguingly, 
it diminishes when the privacy level surpasses a specific threshold.
Furthermore, 
our analysis shows that gradient clipping within DP-SGD can effectively postpone the growth of this gap with increasing privacy levels. 
This study provides critical insights into the complex interplay between privacy and fairness in AI systems. 
It paves the way for future research to advance the understanding of this vital relationship.


\def\refname{REFERENCES}

\end{document}